\documentclass{article}

 \usepackage[preprint]{neurips_2026}

\AtBeginDocument{\raggedbottom}

\makeatletter
\renewcommand{\@noticestring}{%
  Preprint. \dag~Equal contribution.\quad\S~Corresponding authors.%
}
\makeatother

\renewcommand{\acksection}{\section*{Acknowledgments and Funding}}

\usepackage[utf8]{inputenc} 
\usepackage[T1]{fontenc}    
\usepackage{url}            
\usepackage{booktabs}       
\usepackage{array}          
\usepackage{longtable}      
\usepackage{amsfonts}       
\usepackage{nicefrac}       
\usepackage{microtype}      
\usepackage{xcolor}         
\usepackage{colortbl}       
\definecolor{cinemrrow}{RGB}{226,238,250} 
\usepackage{amsmath}        
\usepackage{enumitem}       
\usepackage{algorithm}
\usepackage{algorithmicx}
\usepackage{algcompatible}
\usepackage{algpseudocode}
\usepackage{multirow}
\usepackage{placeins}
\usepackage{subcaption}

\definecolor{cinemrheart}{RGB}{197,48,48} 
\definecolor{mydarkblue}{rgb}{0, 0.08, 0.45}

\usepackage{graphicx}
\usepackage{pifont}
\usepackage{hyperref}
\hypersetup{
  colorlinks=true,
  citecolor=mydarkblue,
  linkcolor=mydarkblue,
  urlcolor=mydarkblue,
  breaklinks=true
}

\title{CineMR: Tool-Integrated Vision-Language Reasoning for Quantitative Cardiac MRI Assessment}

\author{
\textbf{Kunyang Li}\textsuperscript{1$\dagger$} \quad
\textbf{Hai Nguyen}\textsuperscript{1,2$\dagger$} \quad
\textbf{Joshua Lowe}\textsuperscript{1$\dagger$} \quad
\textbf{Chenguang Zhao}\textsuperscript{3} \quad
\textbf{Peace C. Madueme}\textsuperscript{3} \\
\textbf{Mehdi Hedjazi Moghari}\textsuperscript{4} \quad
\textbf{Mubarak Shah}\textsuperscript{1\S} \quad
\textbf{Pegah Khosravi}\textsuperscript{1,2\S} \quad
\textbf{Yuzhang Shang}\textsuperscript{1\S} \\[0.5em]
\textsuperscript{1}Institute for Artificial Intelligence, University of Central Florida \\
\textsuperscript{2}Department of Clinical Sciences, College of Medicine, University of Central Florida \\
\textsuperscript{3}Nemours Cardiac Center, Nemours Children’s Hospital, Florida \\
\textsuperscript{4}Children’s Heart Center, WVU Golisano Children’s, West Virginia
}

\begin{document}

\maketitle

\begin{abstract}
Cardiovascular magnetic resonance (CMR), including cine imaging, is a reference standard for the noninvasive assessment of cardiac morphology and ventricular function. Cine CMR interpretation integrates qualitative visual assessment with quantitative measurements of ventricular volumes, ejection fraction, myocardial mass, wall thickness, and regional wall motion. Current medical vision-language models (VLMs) cannot reliably derive quantitative measurements from multidimensional cine images without analysis tools.
We present \textbf{CineMR}, a tool-augmented VLM that invokes cardiac image-analysis tools and integrates their outputs into interleaved reasoning for quantitative CMR assessment. We also construct a multi-cohort visual question answering benchmark covering quantitative metric extraction, multiclass diagnosis, and differential diagnosis, together with tools for segmentation, phase selection, volumetry, morphometry, and regional wall motion analysis.
CineMR is trained with supervised fine-tuning (SFT) on tool-interaction traces followed by Group Relative Policy Optimization (GRPO) with conditional tool-use rewards.
On the multi-cohort cine CMR benchmark, CineMR achieves \textbf{35.9\% pass@1} and \textbf{58.9\% pass@4}, compared with \textbf{1.5\% pass@1} for the Qwen3-VL-8B backbone and \textbf{0.0\%} and \textbf{7.0\% pass@1} for LLaVA-Med v1.5 and MedGemma-4B, respectively. Correct tool invocation reaches \textbf{99.8\%} after GRPO, up from \textbf{78.9\%} after SFT. Live tool outputs improve ventricular measurement accuracy by \textbf{20.4--23.7\%} over direct model predictions, and removing all tools reduces pass@1 from \textbf{35.9\%} to \textbf{27.9\%}. These results highlight the importance of reliable tool use for quantitative cine CMR reasoning and support CineMR as a promising approach for assistive cardiac image assessment.
Code, benchmark resources, and model weights are available at \url{https://github.com/AI-MIND-Lab/CineMR}.
\end{abstract}

\begin{figure}[t]
  \centering
  \includegraphics[width=\linewidth]{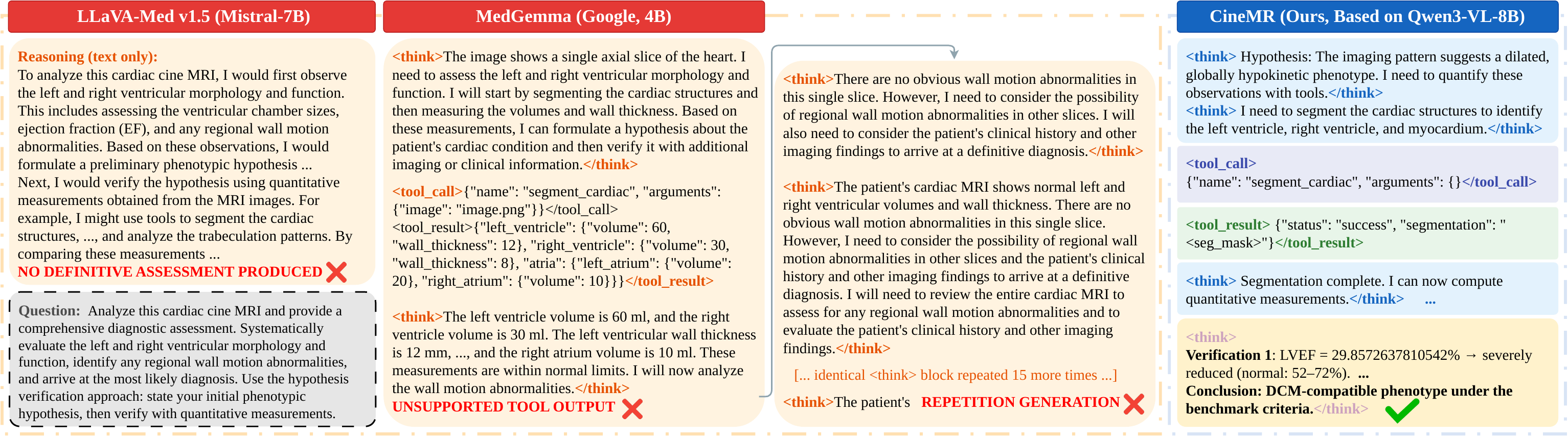}
  \vspace{-1.em}
  \caption{Comparison of model outputs for the same cine CMR case, where the benchmark reference category is dilated cardiomyopathy (DCM). In this illustrative example, LLaVA-Med fails to provide a definitive assessment, whereas MedGemma produces unsupported tool-call content and repetitive text. In contrast, CineMR invokes executable tools to obtain quantitative measurements and evaluates them according to criteria encoded in the study-specific Rule-Based Cardiac Phenotype Classifier, thereby producing a traceable output grounded in explicit measurements and decision rules.}
  \vspace{-0.8em}
  \label{fig:pipeline-base}
\end{figure}

\section{Introduction}
\label{sec:intro}
Cardiovascular magnetic resonance (CMR) is a reference standard for the noninvasive quantification of ventricular volumes, myocardial mass, and systolic function~\citep{petersen2019uk,schulzmenger2020standardized}. Clinical CMR interpretation integrates visual assessment with quantitative measurements of ventricular volumes, ejection fraction, myocardial mass, wall thickness, and regional wall motion, together with clinical history and complementary imaging findings~\citep{esc2023cm, schulzmenger2020standardized}. Quantitative criteria can support phenotype classification, but definitive diagnosis may also require clinical, electrocardiographic, genetic, coronary, and tissue-characterization information.
The combination of multidimensional image analysis, quantitative measurement, and criterion-based interpretation makes CMR a compelling testbed for tool-augmented multimodal reasoning. This is particularly important because errors in upstream measurements can propagate to downstream classifications.

Recent medical VLMs have expanded multimodal question answering and generated-rationale capabilities, with reinforcement learning to optimize structured response generation~\citep{lai2025medr1, pan2025medvlmr1, su2025gmaivlr1, li2023llavamed, sun2025stepbystep}.
However, general-purpose medical VLMs are not designed to perform the multidimensional image analysis required to derive quantitative CMR measurements reliably. As illustrated in Fig.~\ref{fig:pipeline-base}, the evaluated baseline models may fail to produce a definitive assessment or generate repetitive text, unsupported tool outputs, or ungrounded quantitative values when prompted to interpret cine CMR. This limitation reflects the explicit multidimensional computations required for quantitative CMR assessment. For example, assessment of a DCM-compatible phenotype requires delineation of the left ventricle across multiple slices and cardiac phases, voxel-based calculation of ventricular volumes and ejection fraction, and interpretation using reference ranges and study criteria~\citep{petersen2017reference}. Current VLMs cannot be assumed to reproduce this multistep quantitative pipeline reliably from image inputs alone.

Tool-augmented medical agents~\citep{li2024mmedagent, fathi2025aura, wang2025medagentpro, jiang2025ophiuchus} offer a promising direction, although many existing systems emphasize perceptual operations, such as segmentation visualization and region enlargement, rather than the explicit computation of clinically interpretable imaging measurements.
Quantitative descriptors such as the septal-to-lateral wall-thickness ratio and mass-to-volume ratio may contribute to differentiating hypertrophic cardiomyopathy (HCM) from hypertensive heart disease (HHD), but they must be interpreted with additional morphological and clinical information; assessment of a DCM-compatible phenotype incorporates ventricular dilation and systolic dysfunction relative to appropriate sex-specific reference ranges~\citep{petersen2017reference}. These measurements cannot be quantified reliably through visual inspection alone.
These tasks require quantitative measurement in addition to visual interpretation, motivating tools that return explicit quantitative outputs. CineMR addresses this need by equipping a VLM with cardiac image-analysis tools and training it to invoke and integrate their outputs within an interleaved reasoning process.

\noindent \textbf{Rule-Based Cardiac Phenotype Classifier-structured dataset.}
To make the supervision clinically traceable and systematically cover criterion-based reasoning, we organize the benchmark around a Rule-Based Cardiac Phenotype Classifier rather than relying only on direct image-to-label associations. The Rule-Based Cardiac Phenotype Classifier provides explicit diagnostic pathways linking quantitative measurements to reference answers and expected tool use. We derive the training data from a study-specific Rule-Based Cardiac Phenotype Classifier informed by published cardiomyopathy criteria, CMR reference studies, and explicitly documented project-specific thresholds~\citep{esc2023cm, petersen2017reference}. The Rule-Based Cardiac Phenotype Classifier covers ten harmonized outcome categories. Each question maps to a Rule-Based Cardiac Phenotype Classifier node annotated with its task-complexity level (L1–L6) and the tools expected to generate the reference answer. This structure encourages the model to follow criterion-based pathways and provides traceable supervision for both answer generation and tool use.
\begin{figure}[t]
  \centering
  \includegraphics[width=\linewidth]{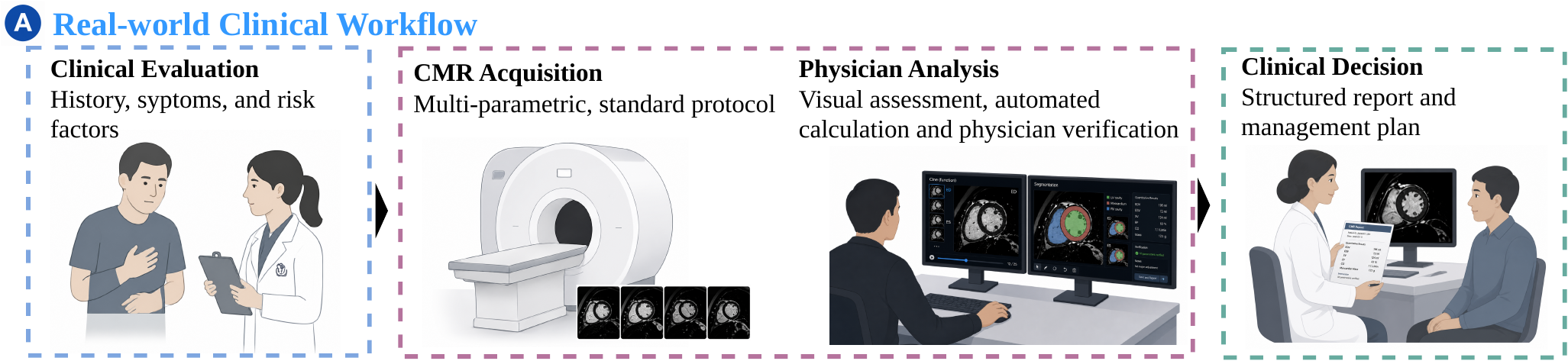}%
  \vspace{-0.2em}
  \includegraphics[width=\linewidth]{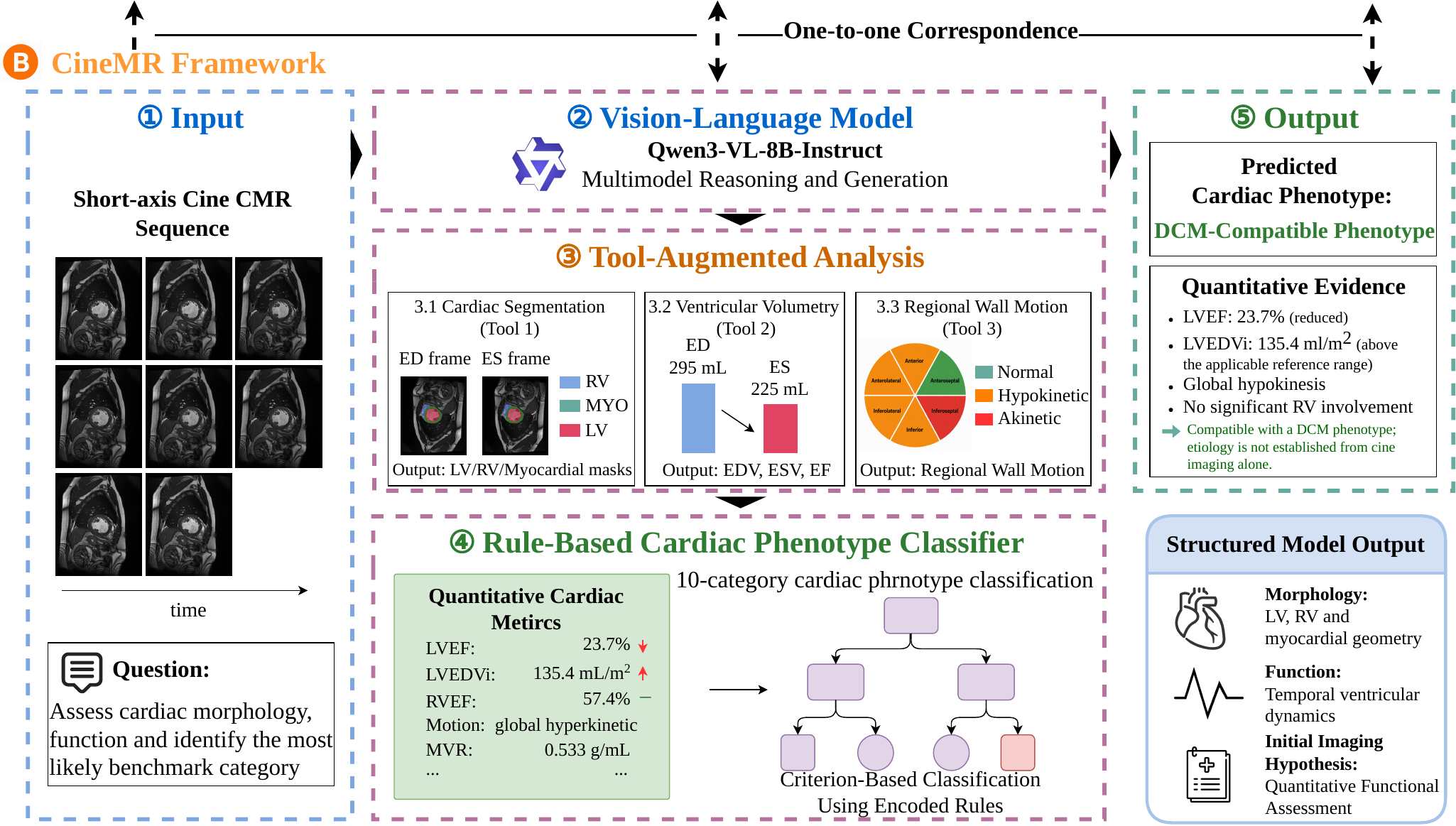}%
  \caption{Overview of the CineMR framework, designed to mirror the real-world cardiologist workflow. Just as a clinician first inspects the cine images to form a visual impression, CineMR generates an initial phenotypic hypothesis via chain-of-thought reasoning. The clinician then orders quantitative analyses (segmentation, volumetry, wall thickness measurement); analogously, CineMR invokes a cardiology-specific quantitative tool suite to obtain precise metrics (EDV, ESV, EF, mass, MVR, wall thickness, sphericity index, regional wall motion). Finally, where the clinician interprets these measurements against ESC/AHA guideline thresholds to reach a diagnosis, CineMR routes them through a guideline-aligned Rule-Based Cardiac Phenotype Classifier with threshold-gated branching, producing an auditable reasoning chain that parallels the clinical diagnostic process from visual assessment to quantitative verification to evidence-based conclusion.}
  \vspace{-0.8em}
  \label{fig:teaser}
\end{figure}
\FloatBarrier

\noindent \textbf{Cardiac image-analysis tool suite.}
We develop a cardiology-specific tool suite for quantitative cine CMR analysis. 
LV, RV, and myocardial segmentation is performed using a task-specific model fine-tuned from CineMA~\citep{fu2025cinema}. 
Voxel-based volumetry computes biventricular end-diastolic volume (EDV), end-systolic volume (ESV), and ejection fraction (EF). 
Morphometric analysis further derives myocardial mass, mass-to-volume ratio (MVR), maximal wall thickness, septal-to-lateral wall-thickness ratio, and sphericity index. 
Regional wall motion is assessed at the segment level using an AHA segment-based representation. 
These tools provide the VLM with segmentation-derived quantitative measurements. 
The measurements are used at the principal quantitative branches of the study-specific Rule-Based Cardiac Phenotype Classifier.
This design allows the model to condition its responses on explicit quantitative evidence rather than on image features alone.

\noindent \textbf{Two-stage training with conditional tool rewards.}
A conventional approach to tool-use training is supervised fine-tuning (SFT) on reference tool-interaction traces. 
SFT teaches the model the tool-call syntax, the interleaved reasoning format, and the association between different question types and their expected tools. 
However, SFT mainly imitates predefined trajectories and does not directly optimize whether the model selects tools appropriately during its own generation. 
We therefore further apply Group Relative Policy Optimization (GRPO)~\citep{shao2024deepseekmath} to refine the model's tool-use policy. 
The GRPO stage uses a conditional tool-use reward, assigning credit for tool invocation only when the final answer is correct. 
This encourages the model to learn when tool use is beneficial while reducing unnecessary or indiscriminate tool calls.

We evaluate CineMR on a multi-cohort benchmark constructed from ACDC~\citep{bernard2018acdc}, M\&Ms, and M\&Ms-2~\citep{campello2021mnms, martin2023mnms2}. The benchmark spans ten harmonized outcome categories and six levels of task complexity (L1–L6). CineMR achieves 35.9\% pass@1 and 58.9\% pass@4 overall, compared with 1.5\% pass@1 (4.5\% pass@4) for the unfine-tuned Qwen3-VL-8B backbone, 7.0\% pass@1 (18.0\% pass@4) for MedGemma-4B-it, and 0.0\% pass@1 (0.0\% pass@4) for LLaVA-Med v1.5. On L3–L4 tasks requiring criterion-based quantitative assessment, CineMR achieves 58.7\% pass@1 and 85.5\% pass@4, whereas end-to-end multicategory diagnosis across all ten benchmark categories (L5) remains substantially harder, at 13.3\% pass@1, indicating that CineMR is not yet a reliable standalone diagnostic system. GRPO also raises correct tool selection to 99.8\% of gold tool-requiring items, up from 78.9\% after supervised fine-tuning, and live tool outputs improve ventricular measurement accuracy by 20.4--23.7\% over direct model predictions, highlighting the importance of reliable tool use for quantitative cine CMR reasoning and supporting tool-augmented VLMs as a promising approach for assistive cardiac image assessment.

\begin{figure}[t]
  \centering
  \includegraphics[width=\linewidth]{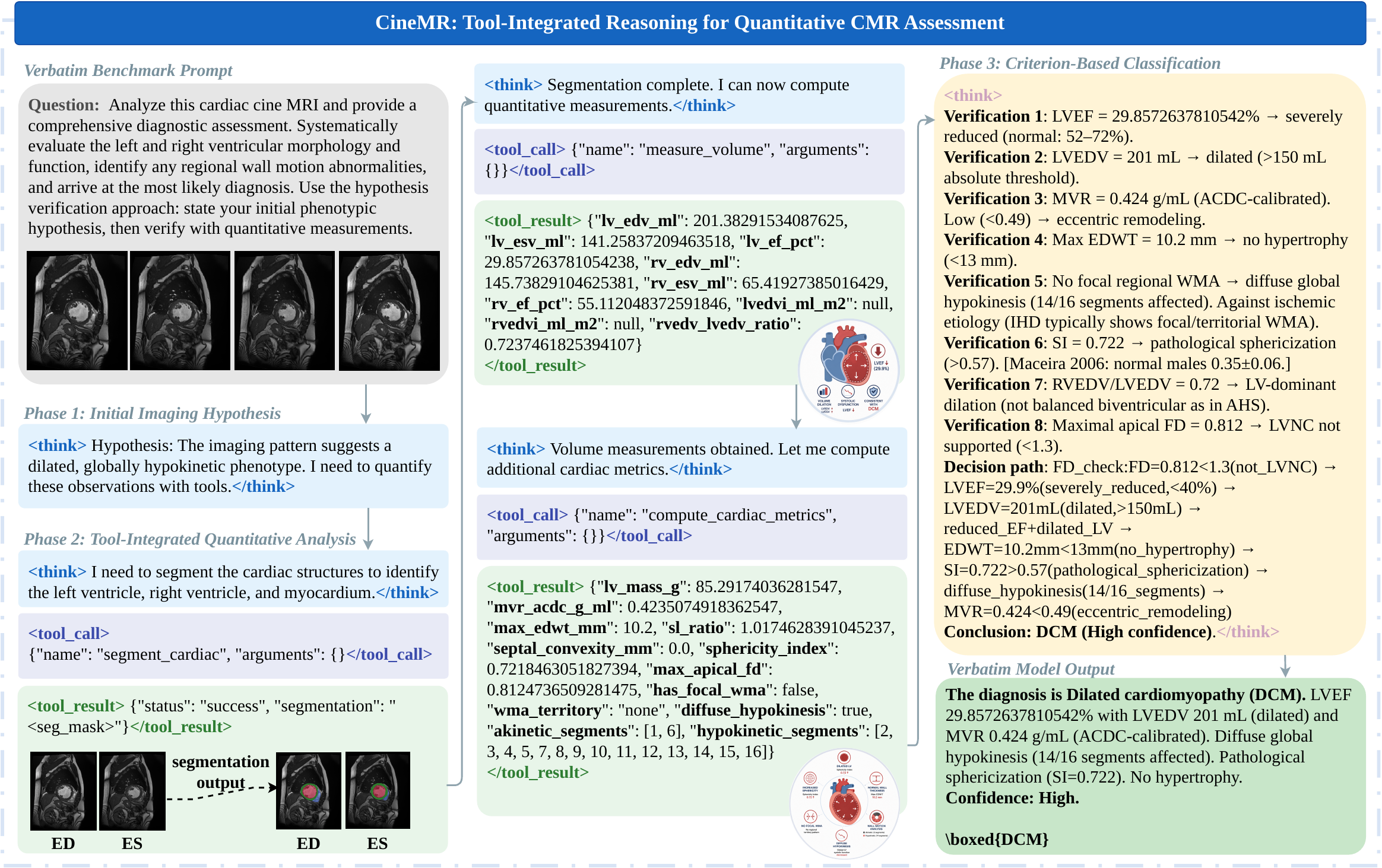}
  \vspace{-1.em}
  \caption{
  Representative verbatim CineMR rollout for a cine CMR case assigned to the DCM category. The figure shows the original benchmark prompt, model-generated tool calls, executable tool outputs, intermediate generated rationale, and final predicted category. CineMR invokes cardiac segmentation, ventricular volumetry, and morphometric analysis tools before applying the published and study-specific criteria encoded in the Rule-Based Cardiac Phenotype Classifier. Tool results are displayed at their native quantitative precision. The intermediate rationale and final response are reproduced verbatim and should not be interpreted as independently validated clinical conclusions. 
  }
  \vspace{-0.8em}
  \label{fig:pipeline-ours}
\end{figure}
\FloatBarrier

\section{Related Work}
\label{sec:related}


\paragraph{Tool-Augmented Medical Agents.}
Integrating multimodal large language models (MLLMs) with external tools can extend their capabilities beyond image and text generation.
MMedAgent~\citep{li2024mmedagent}, VILA-M3~\citep{nath2025vilam3}, and AURA~\citep{fathi2025aura} incorporate task-specific tools for operations such as segmentation, classification, and image annotation.
MedAgent-Pro~\citep{wang2025medagentpro} and AgentMD~\citep{jin2025agentmd} organize specialized models or tools within agent-based diagnostic or clinical risk-assessment workflows.
Ophiuchus~\citep{jiang2025ophiuchus} is particularly relevant because it uses a three-stage training strategy comprising supervised fine-tuning, self-reflection fine-tuning, and agentic Group Relative Policy Optimization~\citep{shao2024deepseekmath} to train a VLM to select and invoke image-analysis tools within an interleaved generation process.
Relative to the requirements of quantitative CMR assessment, many of these approaches emphasize perceptual tools, such as segmentation visualization and region enlargement, rather than tools that return explicit imaging measurements. In addition, several systems rely primarily on prompting or supervised fine-tuning for tool selection. CineMR extends this paradigm by introducing quantitative tools that derive explicit quantitative measurements from segmentation outputs and by organizing training data around a study-specific Rule-Based Cardiac Phenotype Classifier designed to cover the encoded benchmark criteria systematically.

\paragraph{AI for Cardiac MRI Analysis.}
Deep learning methods have shown strong performance on several cine CMR analysis tasks. The ACDC challenge~\citep{bernard2018acdc}, M\&Ms challenge, and M\&Ms-2 challenge~\citep{campello2021mnms, martin2023mnms2} provide widely used benchmarks for cardiac segmentation. Representative segmentation approaches include nnU-Net~\citep{isensee2021nnunet} and task-specific models fine-tuned from the CineMA foundation model~\citep{fu2025cinema}. Prior work has also developed automated pipelines for quality-controlled segmentation and quantitative characterization of cardiac function~\citep{ruijsink2020automated}, as well as radiomics models for characterizing cardiovascular risk-related imaging phenotypes~\citep{cetin2020radiomics}. To our knowledge, prior work has not combined executable quantitative cine CMR analysis tools with an interleaved VLM generation process for criterion-based cardiac phenotype assessment. CineMR differs from these approaches by training a VLM to generate an initial imaging hypothesis, invoke executable tools for quantitative analysis, and integrate the returned measurements using criteria encoded in a study-specific Rule-Based Cardiac Phenotype Classifier.

\section{Method}
\label{sec:method}

CineMR trains a vision-language model to perform criterion-based cardiac phenotype assessment by strategically invoking quantitative tools within an interleaved generation process.
An overview of the proposed framework is shown in Fig.~\ref{fig:teaser}.
We first formulate the diagnostic reasoning problem (\S\ref{sec:formulation}), and then introduce the Rule-Based Cardiac Phenotype Classifier-structured dataset (\S\ref{sec:dataset}), the quantitative tool suite (\S\ref{sec:tools}), and the two-stage training pipeline (\S\ref{sec:training}).

\subsection{Problem Formulation}
\label{sec:formulation}

Let $\mathcal{I}$ denote a cine CMR sequence, $q$ a benchmark question, and $\mathcal{T} = \{t_1, \ldots, t_K\}$ the available tool set. 
We model the task as an agentic generation process: at each turn $i$, the model produces reasoning text $r_i$ or emits a tool invocation $\tau_i = (t_k, a_i)$; the runtime executes $t_k$ on the imaging input I using arguments $a_i$ and returns $\mathbf{o}_i = t_k(I, a_i)$ which is inserted into the generation context. The full response $y = (r_1, \tau_1, \mathbf{o}_1, r_2, \ldots, r_n)$ ends with a predicted answer $\hat{d} \in \mathcal{D}$, where $D$ may represent a numerical value, binary finding, or benchmark category, depending on the question type. 

As shown in Fig.~\ref{fig:pipeline-ours}, this proceeds in three phases: (1) a \texttt{<think>} block forming an initial imaging hypothesis, (2) one or more \texttt{<tool\_call>}/\texttt{<tool\_result>} turns obtaining quantitative measurements, and (3) a narrative synthesis that evaluates the returned measurements using the published and study-specific criteria encoded in the Rule-Based Cardiac Phenotype Classifier before producing the final predicted category $\hat{d}$.

\subsection{Rule-Based Cardiac Phenotype Classifier-Structured Dataset}
\label{sec:dataset}

Published cardiomyopathy guidelines and CMR reference studies contain quantitative criteria that can inform phenotype assessment. We encode a subset of these criteria, together with explicitly documented study-specific thresholds, into benchmark decision rules, including criteria related to LVEF, wall thickness, ventricular volumes, and regional wall motion.
Several CMR findings relevant to cardiac phenotype assessment can be represented using quantitative criteria. However, definitive clinical diagnosis also incorporates history, symptoms, electrocardiography, genetics, loading conditions, coronary assessment, and tissue-characterization imaging.
This design provides three advantages: (i) criterion-traceable supervision based on deterministic answer-generation rules rather than unconstrained LLM-generated labels, (ii) node-level annotations specifying the tools expected by the reference policy, and (iii) six task-complexity levels (L1–L6).

In clinical practice, CMR interpretation integrates visual assessment with quantitative measurements and other available clinical and imaging information rather than exhaustively traversing a fixed Rule-Based Cardiac Phenotype Classifier~\citep{schulzmenger2020standardized}.
The decision rules represent the verification pathways encoded for this benchmark, while the model learns to select tools and integrate their outputs using the corresponding published and study-specific criteria.
For each patient, routing the extracted metrics through the Rule-Based Cardiac Phenotype Classifier yields a candidate diagnosis, which is reconciled against the cohort's independently confirmed clinical label: whenever the two disagree, the confirmed cohort label takes precedence and becomes the boxed reference answer. This reconciliation step overrode the classifier's candidate diagnosis in only 2 of 3,320 patients, indicating that the encoded rules agree with the cohort's confirmed diagnoses in the overwhelming majority of cases.

\begin{figure}[t]
  \centering
  \includegraphics[width=\linewidth]{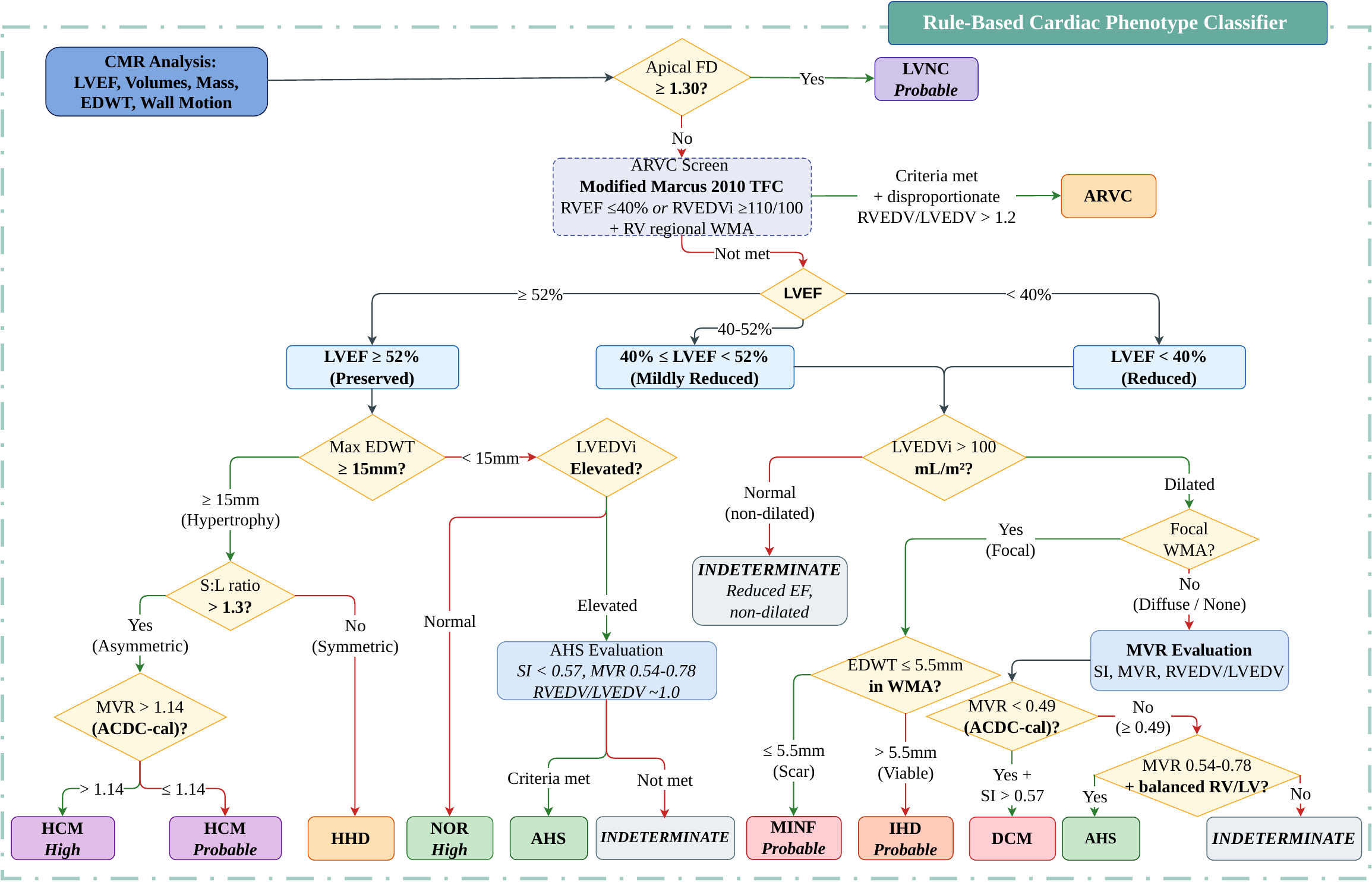}
  \vspace{-1.em}
  \caption{The tools produce the quantitative measurements required by the study-specific Rule-Based Cardiac Phenotype Classifier. Final classification includes the following categories: hypertrophic cardiomyopathy (HCM), athlete’s heart or physiological athletic remodeling, arrhythmogenic right ventricular cardiomyopathy (ARVC), left ventricular non-compaction (LVNC), ischemic heart disease (IHD), myocardial infarction (MINF), dilated cardiomyopathy (DCM), hypertensive heart disease (HHD), normal (NOR), and indeterminate.}
  \vspace{-0.8em}
  \label{fig:tree}
\end{figure}

The Rule-Based Cardiac Phenotype Classifier in Fig.~\ref{fig:tree} assigns cases to ten harmonized benchmark categories: NOR, DCM, HCM, HHD, MINF, IHD, ARVC, athlete’s heart, LVNC, and INDETERMINATE. Each branching point uses either a published quantitative criterion or an explicitly documented project-specific threshold or calibration.
LVEF classification uses sex-specific reference ranges reported by~\citep{petersen2017reference}. The HCM-related benchmark rules are informed by the 2023 ESC cardiomyopathy guideline~\citep{esc2023cm}. In adults, otherwise unexplained maximal LV wall thickness of $\geq$15 mm supports HCM, whereas a threshold of $\geq$13 mm may support HCM in first-degree relatives of patients with unequivocal HCM or in the presence of other compelling clinical or genetic evidence. The ARVC-related benchmark rules use the structural and functional CMR components of the 2010 Revised Task Force Criteria~\citep{marcus2010arvc}. For ACDC, MVR thresholds are adjusted using a project-specific factor of 0.913 to account for differences in the treatment of papillary muscles. This adjustment is informed by studies demonstrating the effects of papillary-muscle inclusion on LV mass and volume measurements~\citep{vogelclaussen2006pure, gommans2016high}. No corresponding adjustment is applied to M\&Ms or M\&Ms-2.
\FloatBarrier

\noindent\textbf{Task-complexity levels.}
We organize questions into six cognitive layers of increasing reasoning complexity:

\begin{itemize}[nosep,leftmargin=1.5em]
\item \textbf{L1, numeric extraction:} direct retrieval of quantitative metrics, e.g., ``What is the LVEF?'' Tool use is required.
\item \textbf{L2, binary finding assessment:} evaluation of a visual or anatomical finding, e.g., “Is the LV dilated?” Tool use is optional under the benchmark reference policy, except for templates that explicitly require quantitative measurements.
\item \textbf{L3, criterion evaluation:} comparison with a published or study-specific threshold, e.g., ``Does EDWT satisfy the HCM criterion of $\geq 15$\,mm?'' Tool use is required.
\item \textbf{L4, pairwise criterion-based classification:} distinction between two benchmark categories, e.g., HCM versus HHD. Multiple tools are required under the reference policy.
\item \textbf{L5, multicategory assessment:} end-to-end prediction across all ten benchmark categories using a hypothesis-and-verification format.
\item \textbf{L6, targeted differential assessment:} distinction between clinically confusable benchmark categories, e.g., DCM versus MINF.
\end{itemize}

The resulting 19 templates comprise 6, 4, 4, 2, 1, and 2 questions for L1, L2, L3, L4, L5, and L6, respectively.(Appendix~\ref{sec:question_templates} lists them all).
These levels progress from quantitative measurement and criterion evaluation to multicategory and targeted differential assessment, with L3–L6 requiring increasingly selective and compositional tool use under the benchmark reference policy.

\noindent\textbf{Metric provenance.}
We audited all 66 classifier thresholds for clinical traceability: 43 were grounded in guidelines or peer-reviewed studies, while 23 were project-specific thresholds with documented derivations and published precedents.
\subsection{Cardiology-specific Quantitative Tool Suite}
\label{sec:tools}
\begin{table}[t]
\centering
\caption{\textbf{Cardiology-specific tool suite.} Each tool maps to Rule-Based Cardiac Phenotype Classifier nodes requiring quantitative thresholds inaccessible to visual inspection.}
\label{tab:tool_coverage}
\vspace{0.3em}
\small
\begin{tabular}{@{}llll@{}}
\toprule
\textbf{Tool} & \textbf{Input $\rightarrow$ Output} & \textbf{Unlocked Metrics} & \textbf{Role} \\
\midrule
\texttt{segment\_cardiac} & $\mathcal{I} \rightarrow \mathbf{s}$ & LV/RV/MYO masks & prerequisite \\
\texttt{measure\_volume} & $\mathbf{s} \rightarrow \mathbf{v}$ & EDV, ESV, EF, EDVi & computational \\
\texttt{compute\_cardiac\_metrics} & $(\mathbf{s}, \mathbf{v}) \rightarrow \mathbf{m}$ & mass, MVR, EDWT, S:L, SI & computational \\
\texttt{wall\_motion\_analysis} & $\mathbf{s} \rightarrow \mathbf{w}$ & segment-level WMA & computational \\
\texttt{select\_cardiac\_phases} & $\mathbf{s} \rightarrow (t_\text{ED}, t_\text{ES})$ & temporal grounding & auxiliary \\
\texttt{zoom\_anatomy} & $(t, \text{target}) \rightarrow \text{ROI}$ & local visual detail & auxiliary \\
\bottomrule
\end{tabular}
\end{table}
\FloatBarrier

Clinical cardiac MRI diagnosis is fundamentally a \textit{measurement-driven} process: ESC/AHA guidelines define diagnostic criteria through quantitative thresholds, such as LVEF $<40\%$ for reduced systolic function and EDWT $\ge 15\,mm$ for HCM~\citep{esc2023cm, ommen2020aha}. Obtaining these measurements from raw imaging data requires multi--step computation, including anatomical delineation, voxel-wise integration, and derived ratio calculation, which current VLMs cannot reliably perform internally. Many prior tool-augmented medical agents~\citep{li2024mmedagent, jiang2025ophiuchus} emphasize perceptual tools, such as segmentation visualization and region enlargement, rather than tools that return explicit quantitative imaging measurements.
We therefore design a \emph{quantitative} tool suite with $K=6$ tools, organized as a dependency chain in which each tool’s output feeds subsequent computations. These tools produce the quantitative measurements required by the study-specific Rule-Based Cardiac Phenotype Classifier.

\noindent\textbf{Segmentation tool.}
\texttt{segment\_cardiac}$(\mathcal{I}) \rightarrow \mathbf{s}$ uses a task-specific segmentation model fine-tuned from CineMA~\citep{fu2025cinema}, a multiview convolutional-transformer masked-autoencoder foundation model pretrained on approximately 15 million cine CMR images from 74,916 UK Biobank studies, to produce frame-wise masks for the left ventricle (LV), right ventricle (RV), and myocardium (MYO). 
This tool is a prerequisite for downstream quantitative analyses, whose measurements are derived using standardized quantitative post-processing rather than unstructured visual estimation.

\noindent\textbf{Volumetric tools.}
\texttt{measure\_volume}$(\mathbf{s}) \rightarrow \mathbf{v}$ computes computes biventricular end-diastolic and end-systolic volumes, ejection fractions, and, when body surface area is available, indexed volumes using voxel-based volumetry across the short-axis stack. When body surface area is unavailable, indexed volumes are not reported. LVEF and indexed ventricular volumes are major quantitative features used by the study-specific Rule-Based Cardiac Phenotype Classifier~\citep{esc2023cm}. LVEF determines the initial branch preserved, mildly reduced, or reduced while indexed volumes inform subsequent category-specific evaluations. Because reliable application of these criteria requires quantitative measurement, visual estimation alone is insufficient~\citep{schulzmenger2020standardized}.

\texttt{select\_cardiac\_phases}$(\mathbf{s}) \rightarrow (t_{\mathrm{ED}}, t_{\mathrm{ES}})$ identifies end-diastolic and end-systolic frames from the LV cavity volume curve. Accurate phase identification is critical because incorrect ED/ES selection introduces systematic volume errors that propagate to all downstream metrics~\citep{schulzmenger2020standardized}. This tool automates the clinically recommended approach of selecting ED as maximum and ES as minimum LV cavity volume.

\noindent\textbf{Morphometric tool.}
\texttt{compute\_cardiac\_metrics}$(\mathbf{s}, \mathbf{v}) \rightarrow \mathbf{m}$ derives advanced diagnostic features, each linked to a specific branch of the diagnostic Rule-Based Cardiac Phenotype Classifier:
\begin{itemize}[nosep,leftmargin=1.5em]
\item \emph{LV myocardial mass ($V_{\mathrm{MYO}} \times 1.05$\,g/mL):} a standard CMR metric for assessing ventricular hypertrophy~\citep{esc2023cm}.
\item \emph{Mass-to-volume ratio (MVR):} captures disproportionate myocardial thickening relative to chamber volume and supports discrimination between HCM and athlete's heart~\citep{czimbalmos2019demanding}.
\item \emph{Maximal end-diastolic wall thickness (EDWT):} using an AHA segment-based representation and used to evaluate the encoded wall-thickness criteria. 
\item \emph{Septal-to-lateral wall-thickness ratio (S:L):} provides a quantitative descriptor of asymmetric septal hypertrophy and is used as one feature in the study-specific HCM-versus-HHD rule~\citep{henry1973asymmetric}. 
\item \emph{Sphericity index (SI):} 
LV geometry and is used in the study-specific Rule-Based Cardiac Phenotype Classifier as a feature associated with spherical ventricular remodeling~\citep{nakamori2017left}.
\end{itemize}

\noindent\textbf{Wall motion tool.}
\texttt{wall\_motion\_analysis}$(\mathbf{s}) \rightarrow \mathbf{w}$ evaluates regional wall motion by comparing endocardial radial contraction and myocardial thickening between ED and ES frames.
Each AHA segment is classified as normal, hypokinetic, or akinetic.
This tool identifies focal versus diffuse wall-motion patterns. A focal territorial abnormality may support an ischemic phenotype, whereas diffuse hypokinesis may be more compatible with a nonischemic dilated phenotype; however, cine wall motion alone does not establish etiology or myocardial scar.

\section{Training Pipeline}
\label{sec:training}
We adopt a two-stage training paradigm inspired by recent agentic training approaches~\citep{jiang2025ophiuchus, shao2024deepseekmath}, with modifications tailored to quantitative (rather than visual) tool use.

\noindent\textbf{Stage 1: Cold-start SFT.}
We fine-tune Qwen3-VL-8B-Instruct~\citep{bai2025qwen3} with LoRA  (rank 16, $\alpha=32$) on the training partition of the VQA benchmark using reference tool-use trajectories, whose displayed tool outputs are ground-truth-derived by default and may optionally be substituted with model-predicted measurements for trace realism; the boxed reference answer is always ground-truth-derived regardless. This stage trains the model to generate the tool-invocation syntax, follow the interleaved trace format, and associate question types with tools specified in the reference trajectories. We then augment this trajectory pool by rejection sampling: rollouts from the resulting policy are retained as additional SFT trajectories only if they yield a correct boxed answer, satisfy a tool-protocol adherence threshold, and, for classification items, are self-consistent across repeated samples.

\noindent\textbf{Stage 2: Agentic GRPO.}
We apply Group Relative Policy Optimization (GRPO)~\citep{shao2024deepseekmath}, which samples a group of $G$ responses $\{o_i\}_{i=1}^G$ from the current policy $\pi_{\theta_{\text{old}}}$ for each prompt $q$ and optimizes:
\begin{equation}
\label{eq:grpo}
\mathcal{J}_{\text{GRPO}}(\theta) = \mathbb{E}_{q,\,\{o_i\}} \left[ \frac{1}{G} \sum_{i=1}^{G} \Big( \min\!\Big( r_i(\theta)\,\hat{A}_i,\; \mathrm{clip}\big(r_i(\theta),\,1{-}\epsilon,\,1{+}\epsilon\big)\,\hat{A}_i \Big) - \beta\, D_{\mathrm{KL}}\big(\pi_\theta \,\|\, \pi_{\mathrm{ref}}\big) \Big) \right]
\end{equation}
where $r_i(\theta) = \pi_\theta(o_i \mid q) / \pi_{\theta_{\text{old}}}(o_i \mid q)$ is the importance ratio, $\hat{A}_i = (S_i - \mu_G) / \sigma_G$ is the group-normalized advantage computed from per-sample rewards $S_i$, and $D_{\mathrm{KL}}\big(\pi_\theta \,\|\, \pi_{\mathrm{ref}}\big)$ is a KL-divergence penalty (estimated with the unbiased low-variance estimator of~\citet{shao2024deepseekmath}) that regularizes the policy toward a fixed reference policy $\pi_{\mathrm{ref}}$, set to the Stage-1 SFT checkpoint and held frozen throughout Stage 2, with coefficient $\beta$. The composite reward is:

\begin{equation}
\label{eq:reward}
S = (1-w_{\text{fmt}})\,S_{\text{ans}} + w_{\text{fmt}}\,S_{\text{format}} + w_{\text{tool}}\,S_{\text{tool}} + w_{\text{live}}\,S_{\text{live}} - w_{\text{rep}}\,P_{\text{rep}}
\end{equation}

where $S_{\text{ans}} \in [0, 1]$ scores answer accuracy (tolerance-scaled for numeric tasks, exact match for classification), $S_{\text{format}} \in [-0.5, 0]$ penalizes malformed outputs, $S_{\text{tool}} \in [-0.1, +0.3]$ is \emph{conditional} on answer correctness (tool bonuses are awarded only when the final answer is correct, which is intended to discourage indiscriminate tool invocation without corresponding answer accuracy), $S_{\text{live}}$ rewards successful live tool execution, and $P_{\text{rep}}$ penalizes degenerate repetition. Default weights are $w_{\text{fmt}}{=}0.15$, $w_{\text{tool}}{=}0.12$, $w_{\text{live}}{=}0.03$, $w_{\text{rep}}{=}0.5$; these are hand-set defaults rather than the product of a systematic hyperparameter sweep, and we did not evaluate sensitivity to them.

\section{Experiments}
\label{sec:experiments}

\subsection{Setup}
\label{sec:exp_setup}

\paragraph{Datasets.}
We construct a multi-cohort cine CMR visual question answering benchmark from ACDC~\citep{bernard2018acdc}, M\&Ms~\citep{campello2021mnms}, and M\&Ms-2~\citep{martin2023mnms2}. The benchmark contains 11,531 question-answer pairs, including 7,020 training, 1,191 validation, and 3,320 test items across ten harmonized outcome categories. Each sample consists of a CMR volume, a benchmark question, a deterministic reference answer derived from the predefined decision rules, and annotations specifying the expected tool-use policy.

For each source cohort, we construct patient-level partitions based on the official dataset splits before generating questions and paraphrases, and subsequently merge the partitions across cohorts. Thus, all images, questions, paraphrases, tool annotations, and derived measurements associated with a given patient are restricted to a single partition, preventing patient-level data leakage. The validation split is used for checkpoint selection and epoch-level monitoring during both training stages. The test split is used only for final evaluation and is never used for model selection or hyperparameter tuning.

\paragraph{Metrics.}
A sampled response receives a score of $1$ if its final boxed answer satisfies the item's reference-answer specification and $0$ otherwise. Numerical answers are evaluated using prespecified item-level tolerances, while categorical and binary answers are evaluated using normalized exact matching. All reported rollouts execute predicted tool calls live through the tool runtime.

We report pass@1 and pass@4 using the standard averaged pass@$k$ estimator of~\citet{chen2021evaluating},
$\text{pass@}k := \mathbb{E}_{\text{items}}\!\left[1-\tbinom{n-c}{k}/\tbinom{n}{k}\right]$,
where $c$ is the number of correct responses among $n$ sampled rollouts for an item. With $n{=}4$, pass@1 is the mean fraction of correct rollouts per item, while pass@4 indicates whether at least one of the four rollouts is correct, averaged across items. 

We further report two tool-use diagnostics (Figure~\ref{fig:tool_call_return}). The \emph{correct tool call rate} is the fraction of gold tool-requiring items for which the predicted tool set exactly matches the gold tool set. The \emph{correct tool return rate} is the fraction of executed gold-matched tool calls whose returned outputs agree with the ground truth within the prescribed tolerance.

\paragraph{Implementation.}
We fine-tune Qwen3-VL-8B-Instruct on tool-integrated reasoning traces using LoRA with rank $r{=}16$, scaling factor $\alpha{=}32$, and dropout $0.05$. SFT uses AdamW in bf16 with a 4,096-token context window. Starting from the resulting checkpoint, we perform tool-integrated GRPO using LoRA with $r{=}64$ and $\alpha{=}128$, $G{=}4$ rollouts per prompt, a maximum response length of 1,024 tokens, and three training epochs. We use asymmetric clipping with $\epsilon_{\mathrm{lo}}{=}0.2$ and $\epsilon_{\mathrm{hi}}{=}0.3$, and set $\beta{=}0.03$ in Eq.~\ref{eq:grpo}.
The reward weights in Eq.~\ref{eq:reward} are $w_{\mathrm{fmt}}{=}0.15$, $w_{\mathrm{tool}}{=}0.12$, $w_{\mathrm{live}}{=}0.03$, and $w_{\mathrm{rep}}{=}0.5$. These values are fixed heuristically and are not tuned using validation or test performance. Training is performed on NVIDIA A100 and H100 GPUs.

We report results for both the SFT checkpoint and the GRPO checkpoint. We compare against the unfine-tuned Qwen3-VL-8B-Instruct backbone~\citep{bai2025qwen3} using the same CineMR tool-aware system prompt, so that this comparison isolates the effect of training from prompting. We additionally compare against the off-the-shelf medical VLMs LLaVA-Med v1.5~\citep{li2023llavamed} and MedGemma-4B-it~\citep{sellergren2026medgemma}. SFT and GRPO are evaluated using four rollouts per item with temperature $0.7$, from which pass@1 and pass@4 are computed. 

\subsection{Results}
\label{sec:exp_results}

\begin{table}[t]
\centering
\caption{
Held-out test performance by cohort (\(n=3{,}320\); ACDC: 696, M\&Ms: 1,846, M\&Ms-2: 778). All methods use \(k=4\) rollouts with temperature 0.7. SFT and GRPO correspond to the matched pre- and post-RL models under the same live-tool protocol. GRPO consistently improves performance over SFT across all cohorts, yielding the best overall results.
}
\label{tab:test_main_v7}
\vspace{0.5em}
\footnotesize
\renewcommand{\arraystretch}{1.05}
\resizebox{\linewidth}{!}{%
\begin{tabular}{@{}l ccc ccc ccc ccc@{}}
\toprule
& \multicolumn{3}{c}{\textbf{ACDC}} & \multicolumn{3}{c}{\textbf{M\&Ms}} & \multicolumn{3}{c}{\textbf{M\&Ms-2}} & \multicolumn{3}{c}{\textbf{Overall}} \\
\cmidrule(lr){2-4}\cmidrule(lr){5-7}\cmidrule(lr){8-10}\cmidrule(lr){11-13}
\textbf{Method} & P@1 & P@4 & Mean & P@1 & P@4 & Mean & P@1 & P@4 & Mean & P@1 & P@4 & Mean \\
\midrule
LLaVA-Med v1.5 & 0.0 & 0.0 & 0.0 & 0.0 & 0.0 & 0.0 & 0.0 & 0.0 & 0.0 & 0.0 & 0.0 & 0.0 \\
MedGemma-4B & 5.2 & 16.1 & 5.5 & 7.7 & 19.1 & 7.4 & 6.7 & 17.0 & 6.3 & 7.0 & 18.0 & 6.7 \\
Qwen3-VL-8B & 1.3 & 4.9 & 1.4 & 1.4 & 4.2 & 1.4 & 1.9 & 4.6 & 1.4 & 1.5 & 4.5 & 1.4 \\
\midrule
\rowcolor{cinemrrow} \textbf{SFT} & 15.2 & 28.3 & 15.6 & 26.5 & 41.6 & 26.3 & 24.2 & 42.0 & 24.4 & 23.6 & 38.9 & 23.6 \\
\midrule
\rowcolor{cinemrrow} \textbf{GRPO} & 33.9 & 58.0 & 34.7 & 36.2 & 58.0 & 35.4 & 36.9 & 62.0 & 37.3 & \textbf{35.9} & \textbf{58.9} & \textbf{35.7} \\
\bottomrule
\end{tabular}%
}
\end{table}

\textbf{Table~\ref{tab:test_main_v7} presents the main results.} GRPO raises pass@1 from 23.6\% (cold-start SFT) to 35.9\% overall, with the largest single-cohort gain on ACDC (15.2\%$\to$33.9\%) and consistent gains on M\&Ms (26.5\%$\to$36.2\%) and M\&Ms-2 (24.2\%$\to$36.9\%). Under the same $k{=}4$-averaged pass@1 protocol used throughout, LLaVA-Med produces no valid diagnoses (0.0\%), MedGemma reaches 7.0\%, and the un-fine-tuned Qwen3-VL-8B backbone reaches 1.5\%; CineMR's pass@1 is roughly $24\times$ higher than the un-fine-tuned backbone it is trained from, and more than $5\times$ higher than the strongest off-the-shelf baseline.

\subsection{Ablations}
\label{sec:exp_ablate}
\textbf{Table~\ref{tab:grpo_layers} breaks down the main results by cognitive layer.} On the clinical-criteria layers (L3--L4), where a measured value must be evaluated against a predefined clinical criterion, GRPO reaches 58.7\% pass@1, compared with 54.5\% for SFT and at most 3.0\% for either off-the-shelf baseline. This large gap suggests that explicit tool-integrated quantitative reasoning is particularly important when assessment depends on measured evidence rather than qualitative visual appearance alone. On the full-diagnosis layers (L5--L6), GRPO (18.6\%) only marginally improves over SFT (18.3\%), consistent with the compounding difficulty of multi-step reasoning, where errors in upstream segmentation or measurement can propagate through subsequent decision steps. Thus, CineMR's gains are not uniform across visual question answering tasks, but are concentrated on tasks that require converting cine MRI into explicit quantitative evidence and evaluating that evidence against clinical criteria.

\begin{table}[t]
\centering
\caption{
Layer-wise performance of SFT (pre-RL) and GRPO (post-RL) with live tool execution, compared with the off-the-shelf MedGemma-4B-it and Qwen3-VL-8B baselines. Results are reported by cognitive layer on the held-out test set (\(n=3{,}320\)). GRPO improves over SFT across most cognitive layers, with the largest gains observed in Layers 2--3. 
}
\label{tab:grpo_layers}
\vspace{0.5em}
\footnotesize
\renewcommand{\arraystretch}{0.85}
\resizebox{\linewidth}{!}{%
\begin{tabular}{@{}l ccc ccc ccc ccc@{}}
\toprule
& \multicolumn{3}{c}{\textbf{MedGemma-4B}} & \multicolumn{3}{c}{\textbf{Qwen3-VL-8B}} & \multicolumn{3}{c}{\textbf{CineMR (SFT)}} & \multicolumn{3}{c}{\textbf{CineMR (GRPO)}} \\
\cmidrule(lr){2-4}\cmidrule(lr){5-7}\cmidrule(lr){8-10}\cmidrule(lr){11-13}
\textbf{Slice} & P@1 & P@4 & Mean & P@1 & P@4 & Mean & P@1 & P@4 & Mean & P@1 & P@4 & Mean \\
\midrule
Layer 1 & 1.9 & 7.5 & 2.2 & 2.9 & 8.3 & 2.6 & \cellcolor{cinemrrow}6.5 & \cellcolor{cinemrrow}20.2 & \cellcolor{cinemrrow}6.4 & \cellcolor{cinemrrow}10.8 & \cellcolor{cinemrrow}31.5 & \cellcolor{cinemrrow}10.1 \\
Layer 2 & 17.6 & 38.8 & 17.0 & 0.5 & 1.7 & 0.4 & \cellcolor{cinemrrow}27.6 & \cellcolor{cinemrrow}33.0 & \cellcolor{cinemrrow}27.3 & \cellcolor{cinemrrow}65.3 & \cellcolor{cinemrrow}92.0 & \cellcolor{cinemrrow}65.3 \\
Layer 3 & 8.4 & 23.3 & 7.2 & 0.6 & 1.9 & 0.6 & \cellcolor{cinemrrow}55.2 & \cellcolor{cinemrrow}83.7 & \cellcolor{cinemrrow}55.8 & \cellcolor{cinemrrow}60.9 & \cellcolor{cinemrrow}86.7 & \cellcolor{cinemrrow}60.7 \\
Layer 4 & 2.8 & 15.0 & 4.2 & 0.0 & 1.9 & 0.5 & \cellcolor{cinemrrow}50.5 & \cellcolor{cinemrrow}78.5 & \cellcolor{cinemrrow}52.1 & \cellcolor{cinemrrow}45.8 & \cellcolor{cinemrrow}78.5 & \cellcolor{cinemrrow}50.0 \\
Layer 5 & 0.0 & 0.0 & 0.0 & 0.0 & 0.0 & 0.0 & \cellcolor{cinemrrow}12.9 & \cellcolor{cinemrrow}33.3 & \cellcolor{cinemrrow}12.6 & \cellcolor{cinemrrow}13.3 & \cellcolor{cinemrrow}32.5 & \cellcolor{cinemrrow}13.0 \\
Layer 6 & 0.0 & 1.5 & 0.4 & 0.0 & 1.5 & 0.4 & \cellcolor{cinemrrow}37.9 & \cellcolor{cinemrrow}50.0 & \cellcolor{cinemrrow}35.6 & \cellcolor{cinemrrow}37.9 & \cellcolor{cinemrrow}37.9 & \cellcolor{cinemrrow}37.1 \\
\midrule
\textbf{All layers} & 7.0 & 18.0 & 6.7 & 1.5 & 4.5 & 1.4 & \cellcolor{cinemrrow}23.6 & \cellcolor{cinemrrow}38.9 & \cellcolor{cinemrrow}23.6 & \cellcolor{cinemrrow}35.9 & \cellcolor{cinemrrow}58.9 & \cellcolor{cinemrrow}35.7 \\
\bottomrule
\end{tabular}%
}
\end{table}
\FloatBarrier

\begin{figure}[!t]
\centering
\begin{subfigure}[t]{0.32\linewidth}
\centering
\includegraphics[width=\linewidth]{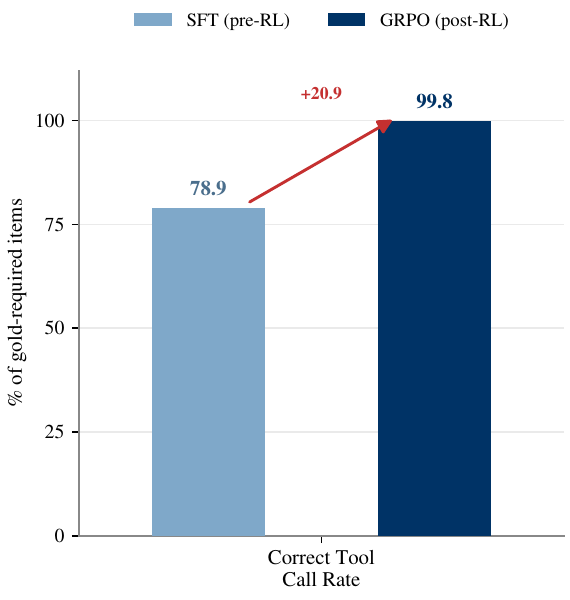}
\caption{Tool call rate.}
\label{fig:tool_call_return}
\end{subfigure}
\hfill
\begin{subfigure}[t]{0.32\linewidth}
\centering
\includegraphics[width=\linewidth]{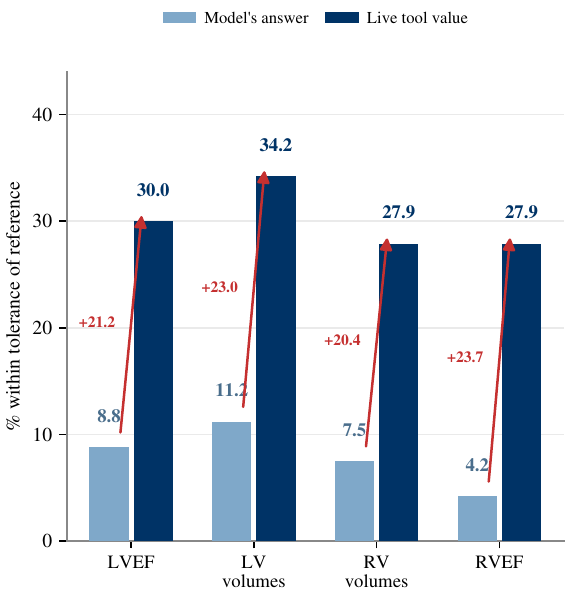}
\caption{Model answer vs.\ live tool value.}
\label{fig:tool_value}
\end{subfigure}
\hfill
\begin{subfigure}[t]{0.32\linewidth}
\centering
\includegraphics[width=\linewidth]{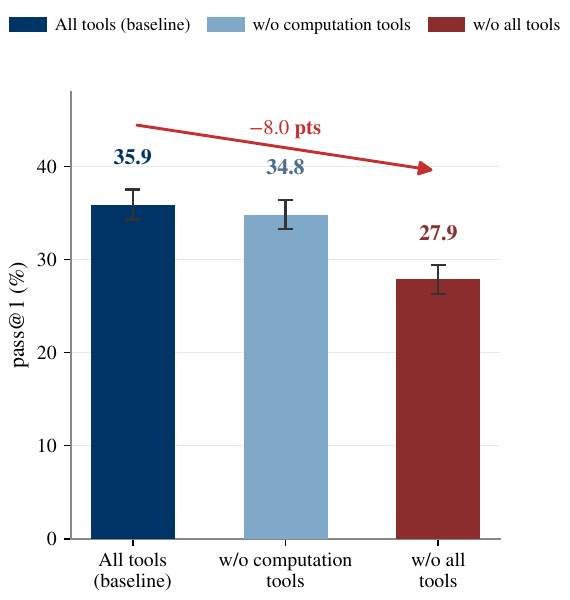}
\caption{Pass@1 with tools withheld.}
\label{fig:ablations}
\end{subfigure}
\caption{
\textbf{Tool-use diagnostics and tool-withholding ablation.}
(a) Correct tool-call rate for SFT and GRPO on gold tool-requiring items.
(b) GRPO accuracy of direct model predictions versus live \texttt{measure\_volume} outputs on ventricular volume and ejection-fraction questions.
(c) GRPO pass@1 on the held-out test set ($n=3{,}320$, $k=4$) with full tool access, without computation tools, and without any tools; error bars show 95\% bootstrap confidence intervals.
GRPO raises correct tool invocation from 78.9\% to 99.8\%; live tool outputs improve measurement accuracy by 20.4--23.7\%, and removing all tools lowers pass@1 from 35.9\% to 27.9\%.
}
\vspace{-0.8em}
\label{fig:tool_diag_ablation}
\end{figure}

Figure~\ref{fig:tool_call_return} further shows that GRPO primarily improves the precision of tool selection. GRPO predicts the exact gold tool set on 99.8\% of gold tool-requiring items, compared with 78.9\% for SFT.
Figure~\ref{fig:tool_value} breaks this down further: across the four volumetric and ejection-fraction question types, the live \texttt{measure\_volume} tool's own returned value is correct substantially more often (28--34\%) than the model's final boxed answer (4--11\%), indicating that part of this gap reflects how the model integrates a returned measurement rather than the measurement's own accuracy.

We next test how much these quantitative outputs contribute to final performance. Withholding the three metric-producing tools at inference time, while retaining segmentation, wall-motion analysis, and visual zoom, reduces pass@1 from 35.9\% to 34.8\% (Figure~\ref{fig:ablations}). Without explicit measurements, the model must instead rely on visual cues and learned priors for questions governed by numerical criteria. Removing all tools, including segmentation, further reduces pass@1 to 27.9\%. The 95\% bootstrap confidence interval for this setting does not overlap with that of the full-tool configuration, indicating that the degradation is unlikely to be attributable to sampling variation alone. Together, these results show that externally computed image-derived evidence contributes information beyond what the VLM recovers from visual reasoning alone.


\FloatBarrier

\section{Conclusion}
\label{sec:conclusion}
We presented CineMR, a tool-augmented vision-language framework for quantitative cine CMR assessment. CineMR combines visual interpretation with dedicated tools for segmentation, phase selection, ventricular volumetry, morphometric measurement, and regional wall-motion analysis, and maps the resulting evidence to benchmark categories using an explicitly encoded Rule-Based Cardiac Phenotype Classifier. On the constructed multi-cohort benchmark, supervised fine-tuning and reinforcement learning substantially improved performance relative to the unfine-tuned backbone, while access to quantitative tools provided an additional, more modest improvement over the tool-ablated configuration. These findings demonstrate the potential of integrating vision-language models with explicit quantitative computation for traceable assessment of cine CMR examinations. However, the benchmark categories and study-specific rules do not constitute complete clinical diagnoses, and the framework has not been validated for autonomous clinical use. Future work should evaluate external generalizability, measurement robustness, calibration, and clinical utility across institutions, scanner vendors, patient populations, and CMR acquisition protocols.

\begin{ack}
This work utilized research computing resources supported by the Advanced Cyberinfrastructure Coordination Ecosystem: Services \& Support (ACCESS) program under allocation ACCESS-CIS240958, including resources provided by the Pittsburgh Supercomputing Center Bridges-2 system. We also thank the AI-MIND Lab for providing GPU resources used in this work. The authors declare no competing interests.
\end{ack}

\bibliographystyle{unsrtnat}
\bibliography{ref}

@article{bernard2018acdc,
  title={{Deep Learning Techniques for Automatic MRI Cardiac Multi-Structures Segmentation and Diagnosis: Is the Problem Solved?}},
  author={Olivier Bernard and Alain Lalande and Cl{\'e}ment Zotti and Fr{\'e}d{\'e}ric Cervenansky and Xin Yang and Pheng-Ann Heng and Irem Cetin and Karim Lekadir and Oscar Camara and Miguel A. gonz{\'a}lez Ballester and Gerard Sanrom{\'a} and Sandy Napel and Steffen Erhard Petersen and Georgios Tziritas and Elias Grinias and Mahendra Khened and Varghese Alex Kollerathu and Ganapathy Krishnamurthi and Marc-Michel Roh{\'e} and Xavier Pennec and Maxime Sermesant and Fabian Isensee and Paul F. J{\"a}ger and Klaus Hermann Maier-Hein and Peter M. Full and Ivo Wolf and Sandy Engelhardt and Christian F. Baumgartner and Lisa M. Koch and Jelmer M. Wolterink and Ivana I{\v{s}}gum and Yeonggul Jang and Yoonmi Hong and Jay Patravali and Shubham Jain and Olivier Humbert and Pierre-Marc Jodoin},
  journal={IEEE Transactions on Medical Imaging},
  year={2018},
  volume={37},
  pages={2514-2525},
  url={https://api.semanticscholar.org/CorpusID:51610194}
}

@article{campello2021mnms,
  title={{Multi-Centre, Multi-Vendor and Multi-Disease Cardiac Segmentation: The M\&Ms Challenge}},
  author={V{\'i}ctor M. Campello and Polyxeni Gkontra and Cristian Izquierdo and Carlos Mart{\'i}n-Isla and Alireza Sojoudi and Peter M. Full and Klaus Hermann Maier-Hein and Yao Zhang and Zhiqiang He and Jun Ma and Mario Parre{\~n}o and Alberto Albiol and Fanwei Kong and Shawn C. Shadden and Jorge Corral Acero and Vaanathi Sundaresan and Mina Saber and Mustafa Elattar and Hongwei Li and Bjoern H Menze and Firas Khader and Christoph Haarburger and Cian M. Scannell and Mitko Veta and Adam Carscadden and K. Punithakumar and Xiao Liu and Sotirios A. Tsaftaris and Xiaoqiong Huang and Xin Yang and Lei Li and Xiahai Zhuang and David Vilad{\'e}s and Martin Descalzo and Andrea Guala and Lucia La Mura and Matthias G. Friedrich and Ria Garg and Julie Lebel and Filipe Henriques and Mahir Karakas and Ersin Cavus and Steffen Erhard Petersen and Sergio Escalera and S. Segu{\'i} and Jos{\'e} F. Rodr{\'i}guez Palomares and Karim Lekadir},
  journal={IEEE Transactions on Medical Imaging},
  year={2021},
  volume={40},
  pages={3543-3554},
  url={https://api.semanticscholar.org/CorpusID:235471536}
}

@article{martin2023mnms2,
  title={{Deep Learning Segmentation of the Right Ventricle in Cardiac MRI: The M\&Ms Challenge}},
  author={Carlos Mart{\'i}n-Isla and V{\'i}ctor M. Campello and Cristian Izquierdo and Kaisar Kushibar and Carla Sendra-Balcells and Polyxeni Gkontra and Alireza Sojoudi and Mitchell J. Fulton and Tewodros Weldebirhan Arega and K. Punithakumar and Lei Li and Xiaowu Sun and Yasmina Al Khalil and Ding Liu and Sana Jabbar and Sandro Queir{\'o}s and Francesco Galati and Moona Mazher and Zheyao Gao and Marcel Beetz and Lennart Tautz and Christoforos Galazis and Marta Varela and Markus Hullebrand and Vicente Grau and Xiahai Zhuang and Domenec Puig and Maria A. Zuluaga and Hassan Mohy-ud-Din and Dimitris N. Metaxas and Marcel M. Breeuwer and Rob J. van der Geest and Michelle L. Noga and St{\'e}phanie Bricq and Mark E. Rentschler and Andrea Guala and Steffen E. Petersen and Sergio Escalera and Jos{\'e} F. Rodr{\'i}guez Palomares and Karim Lekadir},
  journal={IEEE Journal of Biomedical and Health Informatics},
  year={2023},
  volume={27},
  pages={3302-3313},
  url={https://api.semanticscholar.org/CorpusID:258178794}
}

@article{petersen2019uk,
  title={{UK Biobank’s cardiovascular magnetic resonance protocol}},
  author={Steffen Erhard Petersen and Paul M. Matthews and Jane M. Francis and Matthew D. Robson and Filip Zemrak and Redha Boubertakh and Alistair A. Young and Sarah Hudson and Peter J. Weale and Steve Garratt and Rory Collins and Stefan K. Piechnik and Stefan Neubauer},
  journal={Journal of Cardiovascular Magnetic Resonance},
  year={2016},
  volume={18},
  url={https://api.semanticscholar.org/CorpusID:13942043}
}

@article{petersen2017reference,
  title={{Reference ranges for cardiac structure and function using cardiovascular magnetic resonance (CMR) in Caucasians from the UK Biobank population cohort}},
  author={Steffen Erhard Petersen and Nay Linn Aung and Mihir M. Sanghvi and Filip Zemrak and Kenneth Fung and Jos{\'e} Miguel Paiva and Jane M. Francis and Mohammed Yunus Khanji and Elena Lukaschuk and Aaron Mark Lee and Valentina Carapella and Young Jin Kim and Paul Leeson and Stefan K. Piechnik and Stefan Neubauer},
  journal={Journal of Cardiovascular Magnetic Resonance},
  year={2017},
  volume={19},
  url={https://api.semanticscholar.org/CorpusID:4240471}
}

@article{fu2025cinema,
  title={{CineMA: A versatile foundation model for cine cardiac magnetic resonance image analysis tasks}},
  author={Yunguan Fu and Wenjia Bai and Weixi Yi and Charlotte Manisty and Anish N. Bhuva and Thomas A. Treibel and James C. Moon and Matthew J. Clarkson and Rhodri H. Davies and Yipeng Hu},
  journal={arXiv preprint arXiv:2506.00679},
  year={2025}
}

@article{isensee2021nnunet,
  title={{nnU-Net: a self-configuring method for deep learning-based biomedical image segmentation}},
  author={Fabian Isensee and Paul F. Jaeger and Simon A. A. Kohl and Jens Petersen and Klaus Hermann Maier-Hein},
  journal={Nature Methods},
  year={2021},
  volume={18},
  pages={203 - 211},
  url={https://api.semanticscholar.org/CorpusID:227947847}
}

@article{ruijsink2020automated,
  title={{Fully Automated, Quality-Controlled Cardiac Analysis From CMR: Validation and Large-Scale Application to Characterize Cardiac Function}},
  author={Bram Ruijsink and Esther Puyol-Ant{\'o}n and Ilkay Oksuz and Matthew Sinclair and Wenjia Bai and Julia Anne Schnabel and Reza Razavi and Andrew P. King},
  journal={JACC: Cardiovascular Imaging},
  year={2020},
  volume={13},
  pages={684 - 695},
  url={https://api.semanticscholar.org/CorpusID:198134360}
}

@article{cetin2020radiomics,
  title={{Radiomics Signatures of Cardiovascular Risk Factors in Cardiac MRI: Results From the UK Biobank}},
  author={Irem Cetin and Zahra Raisi-Estabragh and Steffen Erhard Petersen and Sandy Napel and Stefan K. Piechnik and Stefan Neubauer and Miguel A. gonz{\'a}lez Ballester and Oscar Camara and Karim Lekadir},
  journal={Frontiers in Cardiovascular Medicine},
  year={2020},
  volume={7},
  url={https://api.semanticscholar.org/CorpusID:226222975}
}

@article{esc2023cm,
  title={{2023 ESC Guidelines for the management of cardiomyopathies.}},
  author={Elena Arbelo and Alexandros Protonotarios and Juan Ram{\'o}n Gimeno and Eloisa Arbustini and Roberto Barriales-Villa and Cristina Basso and Connie R. Bezzina and Elena Biagini and Nico A Blom and Rudolf A. de Boer and Tim De Winter and Perry M. Elliott and Marcus D Flather and Pablo Garc{\'i}a-Pav{\'i}a and Kristina Hermann Haugaa and Jodie Ingles and Ruxandra Oana Jurcut and Sabine Klaassen and Giuseppe Limongelli and Bart L. Loeys and Jens Mogensen and Iacopo Olivotto and Antonis Pantazis and Sanjay Sharma and Jan Peter van Tintelen and James S. Ware and Juan Pablo Kaski and Philippe Charron and Massimo Imazio and Magdy Abdelhamid and Victor Aboyans and Michael Arad and Folkert W. Asselbergs and Riccardo Asteggiano and Zofia Teresa Bilińska and Damien Bonnet and Henning Bundgaard and Nuno Cardim and Jelena {\v{C}}elutkienė and Maja {\v{C}}ike{\v{s}} and Gaetano Maria de Ferrari and Veronica Dusi and Volkmar Falk and Laurent Fauchier and Estelle Gandjbakhch and Tiina Heli{\"o} and Konstantinos C. Koskinas and Dipak Kotecha and Ulf Landmesser and George Lazaros and Basil S. Lewis and A. M. Linhart and Maja-Lisa L{\o}chen and Benjamin Meder and Richard Mindham and James C Moon and Jens Cosedis Nielsen and Steffen E Petersen and Eva Bossano Prescott and Mary N. Sheppard and Gianfranco Sinagra and Marta Sitges and Jacob Tfelt‐Hansen and Rhian M. Touyz and Rogier J. A. Veltrop and Josef Veselka and Karim Wahbi and Arthur A. M. Wilde and Katja Zeppenfeld and Brahim Kichou and Hamayak S. Sisakian and Daniel Scherr and Bernhard L. Gerber and Alen D{\v{z}}ubur and Mariana Gospodinova and Ivo Planinc and Hera Heracleous Moustra and David Zem{\'a}nek and Morten Steen Kvistholm Jensen and Ahmad Samir and Kairit Palm and Eric Schulze-Bahr and Vlachopoulos Haralambos and R{\'o}bert Sepp and Berglind Hl{\'i}n Aðalsteinsd{\'o}ttir and Deirdre Ward and Miry Blich and Afrim Poniku and Olga Lunegova and Ainars Rudzitis and Roland Kassab and Jūratė Barysienė and Steve Huijnen and T. P. De Felice and Eleonora Vataman and Nikola Pavlovi{\'c} and Nawal Doghmi and Folkert W. Asselbergs and Elizabeta Srbinovska Kostovska and Vibeke M Almaas and Elżbieta Katarzyna Biernacka and Dulce Brito and Monica Roșca and Marco Zavatta and Arsen Risti{\'c} and Eva Goncalvesova and Matja{\v{z}} {\v{S}}inkovec and Victoria Ca{\~n}adas-Godoy and Pyotr G. Platonov and Ardan Muammer Saguner and Ahmad Al Saadi and Ikram Kammoun and Ahmet Çelik and E. G. Nesukay and Timur A. Abdullaev and Stefan James and Elena Arbelo and Colin Baigent and Michael A Borger and Sergio Buccheri and Borja Ib{\'a}{\~n}ez and Lars K{\o}ber and Konstantinos C. Koskinas and John Mcevoy and Borislava Mihaylova and Lis Neubeck and Agn{\`e}s Pasquet and Amina Rakisheva and Bianca Rocca and Xavier Rossell{\'o} and Ilonca Vaartjes and Christiaan J M Vrints and Adam Witkowski},
  journal={European Heart Journal},
  year={2023},
  url={https://api.semanticscholar.org/CorpusID:261121505}
}

@article{li2023llavamed,
  title={{LLaVA-Med: Training a Large Language-and-Vision Assistant for Biomedicine in One Day}},
  author={Chunyuan Li and Cliff Wong and Sheng Zhang and Naoto Usuyama and Haotian Liu and Jianwei Yang and Tristan Naumann and Hoifung Poon and Jianfeng Gao},
  journal={ArXiv},
  year={2023},
  volume={abs/2306.00890},
  url={https://api.semanticscholar.org/CorpusID:258999820}
}

@inproceedings{sun2025stepbystep,
  title={{Chiron-o1: Igniting Multimodal Large Language Models towards Generalizable Medical Reasoning via Mentor-Intern Collaborative Search}},
  author={Haoran Sun and Yankai Jiang and Wenjie Lou and Yujie Zhang and Wenjie Li and Lilong Wang and Mianxin Liu and Lei Liu and Xiaosong Wang},
  booktitle={Advances in Neural Information Processing Systems (NeurIPS)},
  year={2025}
}

@article{lai2025medr1,
  title={{Med-R1: Reinforcement Learning for Generalizable Medical Reasoning in Vision-Language Models}},
  author={Yuxiang Lai and Jike Zhong and Ming Li and Shitian Zhao and Xiaofen Yang},
  journal={IEEE Transactions on Medical Imaging},
  year={2025},
  volume={PP},
  url={https://api.semanticscholar.org/CorpusID:277104458}
}

@inproceedings{pan2025medvlmr1,
  title={{MedVLM-R1: Incentivizing Medical Reasoning Capability of Vision-Language Models (VLMs) via Reinforcement Learning}},
  author={Jiazhen Pan and Che Liu and Junde Wu and Fenglin Liu and Jiayuan Zhu and Hongwei Bran Li and Chen Chen and Ouyang Cheng and Daniel Rueckert},
  booktitle={{International Conference on Medical Image Computing and Computer-Assisted Intervention}},
  year={2025},
  url={https://api.semanticscholar.org/CorpusID:276647067}
}

@article{su2025gmaivlr1,
  title={{GMAI-VL-R1: Harnessing Reinforcement Learning for Multimodal Medical Reasoning}},
  author={Yan-Cheng Su and Tian-Xin Li and Jiyao Liu and Chenglong Ma and Junzhi Ning and Cheng Tang and Sibo Ju and Jin Ye and Pengcheng Chen and Ming Hu and Shixiang Tang and Lihao Liu and Bin Fu and Wenqi Shao and Xiaowei Hu and Xiangwen Liao and Yuanfeng Ji and Junjun He},
  journal={ArXiv},
  year={2025},
  volume={abs/2504.01886},
  url={https://api.semanticscholar.org/CorpusID:277502243}
}

@article{shao2024deepseekmath,
  title={{DeepSeekMath: Pushing the Limits of Mathematical Reasoning in Open Language Models}},
  author={Zhihong Shao and Peiyi Wang and Qihao Zhu and Runxin Xu and Jun-Mei Song and Mingchuan Zhang and Y. K. Li and Yu Wu and Daya Guo},
  journal={ArXiv},
  year={2024},
  volume={abs/2402.03300},
  url={https://api.semanticscholar.org/CorpusID:267412607}
}

@article{jiang2025ophiuchus,
  title={{Incentivizing Tool-augmented Thinking with Images for Medical Image Analysis}},
  author={Yankai Jiang and Yujie Zhang and Peng Zhang and Yichen Li and Jintai Chen and Xiaoming Shi and Shihui Zhen},
  journal={ArXiv},
  year={2025},
  volume={abs/2512.14157},
  url={https://api.semanticscholar.org/CorpusID:283909495}
}

@article{li2024mmedagent,
  title={{MMedAgent: Learning to Use Medical Tools with Multi-modal Agent}},
  author={Binxu Li and Tian Yan and Yuanting Pan and Zhe Xu and Jie Luo and Ruiyang Ji and Shilong Liu and Haoyu Dong and Zihao Lin and Yixin Wang},
  journal={ArXiv},
  year={2024},
  volume={abs/2407.02483},
  url={https://api.semanticscholar.org/CorpusID:270877969}
}

@article{nath2025vilam3,
  title={{VILA-M3: Enhancing Vision-Language Models with Medical Expert Knowledge}},
  author={Vishwesh Nath and Wenqi Li and Dong Yang and Andriy Myronenko and Mingxin Zheng and Yao Lu and Zhijian Liu and Hongxu Yin and Yee Man Law and Yucheng Tang and Pengfei Guo and Can Zhao and Ziyue Xu and Yufan He and Greg Heinrich and Stephen R. Aylward and Marc Edgar and Michael Zephyr and Pavlo Molchanov and Baris I Turkbey and Holger R. Roth and Daguang Xu},
  journal={2025 IEEE/CVF Conference on Computer Vision and Pattern Recognition (CVPR)},
  year={2025},
  pages={14788-14798},
  url={https://api.semanticscholar.org/CorpusID:274150020}
}

@article{fathi2025aura,
  title={{AURA: A Multi-Modal Medical Agent for Understanding, Reasoning \& Annotation}},
  author={Nima Fathi and Amar Kumar and Tal Arbel},
  journal={ArXiv},
  year={2025},
  volume={abs/2507.16940},
  url={https://api.semanticscholar.org/CorpusID:280224709}
}

@inproceedings{wang2025medagentpro,
  title={{MedAgent-Pro: Towards Evidence-based Multi-modal Medical Diagnosis via Reasoning Agentic Workflow}},
  author={Ziyue Wang and Junde Wu and Linghan Cai and Chang Han Low and Xihong Yang and Qiaxuan Li and Yueming Jin},
  booktitle={International Conference on Learning Representations (ICLR)},
  year={2026}
}

@article{jin2025agentmd,
  title={{AgentMD: Empowering Language Agents for Risk Prediction with Large-Scale Clinical Tool Learning}},
  author={Qiao Jin and Zhizheng Wang and Yifan Yang and Qingqing Zhu and Donald Wright and Thomas Huang and W. John Wilbur and Zhe He and Andrew Taylor and Qingyu Chen and Zhiyong Lu},
  journal={ArXiv},
  year={2024},
  url={https://api.semanticscholar.org/CorpusID:267759870}
}

@article{marcus2010arvc,
  title={{Diagnosis of Arrhythmogenic Right Ventricular Cardiomyopathy/Dysplasia: Proposed Modification of the Task Force Criteria}},
  author={Frank I. Marcus and William J. McKenna and Duane L. Sherrill and Cristina Basso and Barbara Bauce and David A Bluemke and Hugh Calkins and Domenico Corrado and Moniek G.P.J. Cox and James P. Daubert and Guy Hugues Fontaine and Kathleen Gear and Richard N W Hauer and Andrea Nava and Michael H. Picard and Nikos Protonotarios and Jeffrey E. Saffitz and Danita M. Yoerger Sanborn and Jonathan S. Steinberg and Harikrishna Tandri and Gaetano Thiene and Jeffrey A. Towbin and Adalena Tsatsopoulou and Thomas Wichter and Wojciech Zareba},
  journal={Circulation},
  year={2010},
  volume={121},
  pages={1533-1541},
  url={https://api.semanticscholar.org/CorpusID:15324668}
}

@article{vogelclaussen2006pure,
  title={{Left Ventricular Papillary Muscle Mass: Relationship to Left Ventricular Mass and Volumes by Magnetic Resonance Imaging}},
  author={Jens Vogel-Claussen and John Paul Finn and Antoinette S Gomes and Gregory W. Hundley and Michael Jerosch-Herold and Gregory D. N. Pearson and Shantanu Sinha and Joao A. C. Lima and David A. Bluemke},
  journal={Journal of Computer Assisted Tomography},
  year={2006},
  volume={30},
  pages={426-432},
  url={https://api.semanticscholar.org/CorpusID:39925049}
}

@article{gommans2016high,
  title={{Impact of the papillary muscles on cardiac magnetic resonance image analysis of important left ventricular parameters in hypertrophic cardiomyopathy}},
  author={D.H. Frank Gommans and Jeannette Bakker and Gilbert E Cramer and Freek W. A. Verheugt and Marc A. Brouwer and Marcel J M Kofflard},
  journal={Netherlands Heart Journal},
  year={2016},
  volume={24},
  pages={326 - 331},
  url={https://api.semanticscholar.org/CorpusID:215679534}
}

@article{schulzmenger2020standardized,
  title={{Standardized image interpretation and post-processing in cardiovascular magnetic resonance -- 2020 update: Society for Cardiovascular Magnetic Resonance (SCMR): Board of Trustees Task Force on Standardized Post-Processing}},
  author={Jeanette Schulz-Menger and David A. Bluemke and Jens Bremerich and Scott D. Flamm and Mark A. Fogel and Matthias G. Friedrich and Raymond J. Kim and Florian von Knobelsdorff-Brenkenhoff and Christopher M. Kramer and Dudley J. Pennell and Sven Plein and Eike Nagel},
  journal={Journal of Cardiovascular Magnetic Resonance},
  year={2020},
  volume={22},
  number={1},
  pages={19},
  doi={10.1186/s12968-020-00610-6}
}

@article{ommen2020aha,
  title={{2020 AHA/ACC Guideline for the Diagnosis and Treatment of Patients With Hypertrophic Cardiomyopathy: A Report of the American College of Cardiology/American Heart Association Joint Committee on Clinical Practice Guidelines.}},
  author={Steve R. Ommen and Seema Mital and Michael A. Burke and Sharlene M. Day and Anita Deswal and Perry M. Elliott and Lauren L. Evanovich and Judy W. Hung and Jose A. Joglar and Paul F. Kantor and Carey D. Kimmelstiel and Michelle M. Kittleson and Mark S. Link and Martin S. Maron and Matthew W. Martinez and Christina Y. Miyake and Hartzell V. Schaff and Christopher Semsarian and Paul Sorajja},
  journal={Circulation},
  year={2020},
  url={https://api.semanticscholar.org/CorpusID:227078319}
}

@article{henry1973asymmetric,
  title={{Asymmetric Septal Hypertrophy: Echocardiographic Identification of the Pathognomonic Anatomic Abnormality of IHSS}},
  author={Walter L. Henry and Chester E. Clark and Stephen E. Epstein},
  journal={Circulation},
  year={1973},
  volume={47},
  pages={225–233},
  url={https://api.semanticscholar.org/CorpusID:6722097}
}

@article{nakamori2017left,
  title={{Left ventricular geometry predicts ventricular tachyarrhythmia in patients with left ventricular systolic dysfunction: a comprehensive cardiovascular magnetic resonance study}},
  author={Shiro Nakamori and Haisam Ismail and Long H. Ngo and Warren J. Manning and Reza Vatani Nezafat},
  journal={Journal of Cardiovascular Magnetic Resonance},
  year={2017},
  volume={19},
  url={https://api.semanticscholar.org/CorpusID:28351378}
}

@article{czimbalmos2019demanding,
  title={{The demanding grey zone: Sport indices by cardiac magnetic resonance imaging differentiate hypertrophic cardiomyopathy from athlete’s heart}},
  author={Csilla Czimbalmos and Ibolya Cs{\'e}cs and Attila T{\'o}th and Orsolya Kiss and Ferenc Imre Suhai and N{\'o}ra Syd{\'o} and Zs{\'o}fia Dohy and Astrid Apor and B{\'e}la Merkely and Hajnalka V{\'a}g{\'o}},
  journal={PLoS ONE},
  year={2019},
  volume={14},
  url={https://api.semanticscholar.org/CorpusID:67857631}
}

@article{sellergren2026medgemma,
  title={{MedGemma 1.5 Technical Report}},
  author={Sellergren, Andrew and Gao, Chufan and Mahvar, Fereshteh and Kohlberger, Timo and Jamil, Fayaz and Traverse, Madeleine and Tono, Alberto and Sadjad, Bashir and Yang, Lin and Lau, Charles and others},
  journal={arXiv preprint arXiv:2604.05081},
  year={2026}
}

@article{bai2025qwen3,
  title={{Qwen3-VL Technical Report}},
  author={Bai, Shuai and Cai, Yuxuan and Chen, Ruizhe and Chen, Keqin and Chen, Xionghui and Cheng, Zesen and Deng, Lianghao and Ding, Wei and Gao, Chang and Ge, Chunjiang and others},
  journal={arXiv preprint arXiv:2511.21631},
  year={2025}
}

@article{chen2021evaluating,
  title={{Evaluating Large Language Models Trained on Code}},
  author={Chen, Mark and Tworek, Jerry and Jun, Heewoo and Yuan, Qiming and Pinto, Henrique Ponde de Oliveira and Kaplan, Jared and Edwards, Harri and Burda, Yuri and Joseph, Nicholas and Brockman, Greg and others},
  journal={arXiv preprint arXiv:2107.03374},
  year={2021}
}


\newpage
\appendix

\section{Acronym Glossary}
\label{sec:acronym_glossary}

Table~\ref{tab:acronym_glossary} collects every abbreviation used in the main text and appendix. Each is also spelled out at first use in the running text; this table is provided as a single point of reference.

\begin{table}[h]
\centering
\caption{Acronym glossary.}
\label{tab:acronym_glossary}
\begin{tabular}{ll}
\toprule
\textbf{Acronym} & \textbf{Definition} \\
\midrule
LV / RV & Left / right ventricle (ventricular) \\
LVEF / RVEF & Left / right ventricular ejection fraction \\
LVEDV / LVESV & LV end-diastolic / end-systolic volume \\
RVEDV / RVESV & RV end-diastolic / end-systolic volume \\
EDWT & End-diastolic wall thickness \\
S:L ratio & Septal-to-lateral wall thickness ratio \\
SI & Sphericity index \\
MVR & Mass-to-volume ratio (LV mass / LVEDV) \\
WMA & Wall motion abnormality \\
BSA & Body surface area \\
HCM & Hypertrophic cardiomyopathy \\
HHD & Hypertensive heart disease \\
DCM & Dilated cardiomyopathy \\
ARVC & Arrhythmogenic right ventricular cardiomyopathy \\
MINF & Myocardial infarction (ischemic scar) \\
IHD & Ischemic heart disease \\
AHS & Athlete's heart syndrome \\
LVNC & Left ventricular noncompaction \\
NOR & Normal (no structural/functional abnormality) \\
rTFC & (2010) Revised Task Force Criteria for ARVC \\
ESC / AHA & European Society of Cardiology / American Heart Association \\
SCMR & Society for Cardiovascular Magnetic Resonance \\
CMR & Cardiac magnetic resonance \\
\bottomrule
\end{tabular}
\end{table}

\section{Evaluation Question Templates}
\label{sec:question_templates}

Section~\ref{sec:method} (``Cognitive layer hierarchy'') defines the six layers and gives one illustrative example question per layer. Table~\ref{tab:all_templates} lists the complete set of \textbf{19 question templates} used to construct the ACDC / M\&Ms / M\&Ms-2 benchmark (Section~\ref{sec:exp_setup}), each mapped to its cognitive layer, tool requirement, and the tools it invokes when tools are used. Each template's literal question text below is the \emph{canonical} form; the released dataset additionally includes two LLM-generated paraphrase variants per base question (rephrasing, context enrichment, and specificity variation) for lexical diversity, so a given evaluation item may appear as a paraphrase of, rather than verbatim match to, the text shown here. The canonical templates, their answer-derivation functions, and applicability conditions are implemented in the released code, which is the authoritative source for exact question text, answer formats, and clinical citations.

\newpage

{\footnotesize
\setlength{\tabcolsep}{4pt}
\renewcommand{\arraystretch}{0.95}
\begin{longtable}{@{}>{\raggedright\arraybackslash\ttfamily\footnotesize}p{2.9cm}cc>{\raggedright\arraybackslash}p{8.1cm}@{}}
\caption{All 19 evaluation question templates, by cognitive layer. \textbf{Tools}: whether tool use is required (\checkmark) or optional (--) for that template; a template may override its layer's default (e.g.\ \texttt{L2\_is\_asymmetric} requires tools because the septal:lateral ratio must be computed).}
\label{tab:all_templates}\\
\toprule
\normalfont\textbf{Template ID} & \textbf{Layer} & \textbf{Tools} & \textbf{Question} \\
\midrule
\endfirsthead
\toprule
\normalfont\textbf{Template ID} & \textbf{Layer} & \textbf{Tools} & \textbf{Question} \\
\midrule
\endhead
\midrule \multicolumn{4}{r}{\emph{continued on next page}} \\
\endfoot
\bottomrule
\endlastfoot
L1\_\allowbreak{}lvef & L1 & \checkmark & What is the left ventricular ejection fraction (LVEF) for this patient? Report as a percentage and indicate whether it is normal, mildly reduced, reduced, or severely reduced. \\
L1\_\allowbreak{}lvedv & L1 & \checkmark & Report the left ventricular end-diastolic volume (LVEDV) and end-systolic volume (LVESV) for this cardiac MRI study, including BSA-indexed values if anthropometric data are available. \\
L1\_\allowbreak{}rvedv & L1 & \checkmark & What are the right ventricular end-diastolic volume (RVEDV), end-systolic volume (RVESV), and RV ejection fraction (RVEF) for this patient? \\
L1\_\allowbreak{}lv\_\allowbreak{}mass & L1 & \checkmark & Calculate the left ventricular myocardial mass from this cardiac MRI segmentation. Report the absolute LV mass in grams. \\
L1\_\allowbreak{}rvef & L1 & \checkmark & What is the right ventricular ejection fraction (RVEF) for this patient? Is it within normal limits? \\
L1\_\allowbreak{}max\_\allowbreak{}edwt & L1 & \checkmark & What is the maximum end-diastolic wall thickness (max EDWT) of the left ventricle in this cardiac MRI, and does it exceed the threshold for hypertrophy? \\
L2\_\allowbreak{}is\_\allowbreak{}dilated & L2 & -- & Is the left ventricle dilated in this cardiac MRI? Justify your answer using volumetric measurements. \\
L2\_\allowbreak{}is\_\allowbreak{}hypertrophied & L2 & -- & Does this patient demonstrate left ventricular hypertrophy? Report the maximum wall thickness and indicate whether it is within normal limits. \\
L2\_\allowbreak{}is\_\allowbreak{}asymmetric & L2 & \checkmark & Is the left ventricular hypertrophy asymmetric? Compute the septal-to-lateral wall thickness ratio. \\
L2\_\allowbreak{}has\_\allowbreak{}wma & L2 & -- & Does this cardiac MRI demonstrate regional wall motion abnormalities? If so, which AHA segments are affected and which coronary territory is involved? \\
L3\_\allowbreak{}hcm\_\allowbreak{}threshold & L3 & \checkmark & Does the maximum end-diastolic wall thickness meet the ESC 2023 criterion for hypertrophic cardiomyopathy ($\geq$15\,mm for sporadic cases)? Also assess whether the $\geq$13\,mm familial HCM criterion is met. \\
L3\_\allowbreak{}arvc\_\allowbreak{}criteria & L3 & \checkmark & Do the RV volumetric measurements meet the 2010 Revised Task Force Criteria (rTFC) for ARVC? Assess both major and minor CMR criteria for RVEDVi and RVEF. \\
L3\_\allowbreak{}minf\_\allowbreak{}scar & L3 & \checkmark & In the segment(s) with regional wall motion abnormality, does the end-diastolic wall thickness meet the criterion for probable transmural scar ($\leq$5.5\,mm)? Assess the likelihood of chronic myocardial infarction versus viable hibernating myocardium. \\
L3\_\allowbreak{}dcm\_\allowbreak{}si & L3 & \checkmark & Is the left ventricular sphericity index elevated above the threshold for pathological sphericization ($>$0.57)? Compute the sphericity index from the short-axis stack. \\
L4\_\allowbreak{}hcm\_\allowbreak{}vs\_\allowbreak{}hhd & L4 & \checkmark & This patient has left ventricular hypertrophy with max EDWT $\geq$13\,mm. Is the hypertrophy pattern more consistent with hypertrophic cardiomyopathy (HCM) or hypertensive heart disease (HHD)? Use the S:L ratio, septal convexity, and mass-to-volume ratio in your assessment. \\
L4\_\allowbreak{}dcm\_\allowbreak{}vs\_\allowbreak{}ahs & L4 & \checkmark & The left ventricle is dilated with reduced or borderline ejection fraction and no focal wall motion abnormality. Distinguish between dilated cardiomyopathy (DCM) and athlete's heart syndrome (AHS) using the sphericity index, mass-to-volume ratio, and biventricular volume balance. \\
L5\_\allowbreak{}full\_\allowbreak{}diagnosis & L5 & \checkmark & Analyze this cardiac cine MRI and provide a comprehensive diagnostic assessment. Systematically evaluate the left and right ventricular morphology and function, identify any regional wall motion abnormalities, and arrive at the most likely diagnosis. Use the hypothesis-verification approach: state your initial phenotypic hypothesis, then verify with quantitative measurements. \\
L6\_\allowbreak{}differential\_\allowbreak{}dcm\_\allowbreak{}minf & L6 & \checkmark & This patient has reduced left ventricular ejection fraction and ventricular dilation. The key differential is dilated cardiomyopathy (DCM) versus ischemic heart disease with prior myocardial infarction (MINF). What quantitative features allow you to distinguish between these two diagnoses? \\
L6\_\allowbreak{}differential\_\allowbreak{}hcm\_\allowbreak{}athlete & L6 & \checkmark & This patient is a competitive athlete with increased left ventricular wall thickness (max EDWT in the 13--15\,mm grey zone). Distinguish between hypertrophic cardiomyopathy (HCM) and physiological adaptation (athlete's heart) using mass-to-volume ratio, septal shape, and cavity dimensions. \\
\end{longtable}
}

\section{Limitations and Broader Impact}
\label{sec:limitations}

\paragraph{Segmentation dependency.}
CineMR relies on a dedicated short-axis segmentation stage to derive quantitative cardiac metrics, including ventricular volumes, ejection fractions, myocardial mass, and wall thickness. Because downstream benchmark classification depends on these segmentation-derived measurements, segmentation errors can propagate to the final predicted category. In our error analysis, under- or over-segmentation of the left ventricular cavity was a frequent failure mode in L5 full-assessment tasks and could direct the Rule-Based Cardiac Phenotype Classifier toward an incorrect branch even when the subsequent reasoning process was otherwise consistent with the extracted measurements.

\paragraph{Toward segmentation-free quantitative reasoning.}
As multimodal foundation models and end-to-end learning methods advance, future systems may estimate quantitative cardiac measurements directly from imaging data within an integrated reasoning framework, potentially bypassing explicit segmentation. Such approaches may reduce segmentation-related error propagation; however, their measurement accuracy, robustness, and generalizability across acquisition protocols and scanner vendors will require dedicated validation.

\paragraph{Broader impact.}
By recording tool calls, quantitative outputs, and criterion-based classification paths, CineMR provides more inspectable evidence traces than an end-to-end prediction alone. With prospective external validation, the framework could potentially support quantitative cine CMR assessment in settings where specialist interpretation is limited. However, CineMR is not intended for autonomous clinical use. Its outputs should be reviewed by qualified clinicians, and overreliance on automated results particularly when upstream segmentation is inaccurate could contribute to assessment errors if the system were deployed without appropriate validation, quality control, and safeguards.

\paragraph{Additional limitations.}
This retrospective study uses public research datasets and primarily evaluates short-axis cine CMR. Several benchmark categories incorporate study-specific rules or thresholds and therefore should not be interpreted as complete clinical diagnoses. Future work should include prospective external validation across institutions, scanner vendors, and patient populations and assess calibration, subgroup performance, and robustness to variations in image quality and acquisition protocols.



\end{document}